\documentclass{article}

\usepackage{microtype}
\usepackage{graphicx}
\usepackage{subfigure}
\usepackage{booktabs} 

\usepackage{hyperref}
\usepackage{url}
\usepackage{caption}
\usepackage{times}
\usepackage{latexsym}
\usepackage{graphicx}
\usepackage{tabularx}
\usepackage{url}
\usepackage{array}
\usepackage{setspace}
\usepackage{multirow}
\usepackage{hyperref}
\usepackage{framed}
\usepackage{color}
\usepackage{xcolor}
\usepackage{xspace}
\usepackage{amsmath}
\usepackage{amssymb}
\usepackage{amsthm}
\usepackage{booktabs}
\usepackage{colortbl}
\usepackage[normalem]{ulem}
\usepackage{times}
\usepackage{tikz}
\usepackage[inline]{enumitem}

\usepackage{hyperref}

\newcommand{\spouse}{\textsc{Spouse}}
\newcommand{\parcus}{\textsc{Parcus}}
\newcommand{\moviereview}{\textsc{MovieReview}}

\usepackage[accepted]{icml2020}

\icmltitlerunning{Concept Matching for Low-Resource Classification}

\begin{document}

\newcommand{\norm}[1]{\left\lVert#1\right\rVert}
\newcommand{\ie}{i.e., }
\newcommand{\eg}{e.g., }
\newcommand{\quotes}[1]{``#1''}

\newtheorem{proposition}{Proposition}[section]

\twocolumn[
\icmltitle{Concept Matching for Low-Resource Classification}



\begin{icmlauthorlist}
\icmlauthor{Federico Errica}{pisa}
\icmlauthor{Ludovic Denoyer}{fair}
\icmlauthor{Bora Edizel}{fair}
\icmlauthor{Fabio Petroni}{fair} \\
\icmlauthor{Vassilis Plachouras}{fair}
\icmlauthor{Fabrizio Silvestri}{fair}
\icmlauthor{Sebastian Riedel}{fair,ucl}
\end{icmlauthorlist}

\icmlaffiliation{pisa}{Department of Computer Science, University of Pisa, Pisa, Italy}
\icmlaffiliation{fair}{Facebook AI}
\icmlaffiliation{ucl}{Department of Computer Science, University College London, London, United Kingdom}
\icmlcorrespondingauthor{Federico Errica}{federico.errica@phd.unipi.it}

\icmlkeywords{low resource classification, prototypes, natural language processing}

\vskip 0.3in
]



\printAffiliationsAndNotice{}  

\begin{abstract}
We propose a model to tackle classification tasks in the presence of very little training data. 
To this aim, we approximate the notion of exact match with a theoretically sound mechanism that computes a probability of matching in the input space. Importantly, the model learns to focus on elements of the input that are relevant for the task at hand;
by leveraging highlighted portions of the training data, an error boosting technique guides the learning process. In practice, it increases the error associated with relevant parts of the input by a given factor. Remarkable results on text classification tasks confirm the benefits of the proposed approach in both balanced and unbalanced cases, thus being of practical use when labeling new examples is expensive. In addition, by inspecting its weights, it is often possible to gather insights on what the model has learned.

\end{abstract}

\section{Introduction}
\label{introduction}
Gathering and labeling data is a task that can be expensive in terms of time, human effort and resources.  When practitioners cannot rely on public large datasets, training a model with acceptable performance on a few data points becomes critical in a variety of applications. It is not uncommon that the data is also imbalanced, and as such the demands of gathering samples of the minority class are high. A natural domain in which these issues arise is, for instance, text classification, with notable tasks being hate-speech \cite{waseem-hateful} 
and abuse detection \cite{abuse-detection}. For these reasons, the study of techniques that address this problem can have a tangible impact on society.

One effective approach to overcome the lack of training data is to augment the elements of the input with extra annotations, which has proved to be effective when coupled with feature engineering approaches \cite{rational-improve-text-zaidan2007,modeling-annotators-generative}. Such annotations, \eg highlighted words in a sentence, serve the purposes of guiding the learning process toward good solutions and to prevent overfitting the scarce amount of training samples. The goal of this work is to investigate this idea from a pure representation learning perspective, where there is no human intervention on the raw data but for the extra annotations.

To tackle this challenge, we design an architecture that learns to extract relevant \textit{semantic} concepts from each input sample, such as words in a sentence or nodes in a graph. We assume each input is made by a set of individual representations: in scenarios like natural language processing where words are the main constituents of the input, we can rely on unsupervised pre-trained methods to represent them as vectors \cite{fasttext, bert,pretraining-gnns}.  As we act solely on the model, the technique is flexible and task-agnostic; this is in contrast with task-dependent feature engineering methods \cite{modeling-annotators-generative}. Here, the task is assumed to be new, and as such labels need be (slowly) gathered by someone with domain-specific expertise.

In particular, we introduce a new mechanism to match concepts in each input sample and an effective error \quotes{boosting} technique to exploit the additional annotations. We also provide a theoretical analysis that justifies the choice of our matching mechanism; on the empirical side, we will see how cheap annotation costs can make up for a much larger number of training samples, that is a \textit{desiderata} for low-resource classification. \\ Additionally, in this scenario, it is of practical importance to have some degree of reassurance about what the model has learned; by direct inspection of the weights, we show how it is possible to gain human-readable insights about its decision process. Results across a consistent number of baselines indicate a significant improvement in performance with respect to neural competitors as well as foundational methods that make use of the given annotations.

To summarize, we make the following contributions:
\begin{enumerate*}[label=(\roman*)] 
\item We introduce \parcus{}, a new architecture that effectively combines concept matching and error boosting techniques for low-resource classification;
\item We support our intuition with a theoretical analysis;
\item We empirically validate the approach against a consistent number of baselines, demonstrating strong performance improvements;
\item We perform ablation studies to disentangle the contributions of the architecture main constituents;
\item Qualitative analyses show that the model works according to intuition and can be inspected to gain insights into what it has learned
\end{enumerate*}.

The rest of the paper is structured as follows: Section \ref{related-works} reviews the existing literature; Section \ref{model} formally introduces our model; Section \ref{experiments} details our experiments and discusses our findings; Section \ref{sec:limitations} analyzes limitations future works; finally, Section \ref{conclusion} summarizes our work.

\section{Related Works}
\label{related-works}
There are different ways in which extra annotations can be used. Some works \textit{generate} annotations as a way to interpret the model, while others \textit{exploit} them to inform the learning process. Natural language processing is the field in which these techniques have been investigated the most. 
In particular, the method proposed by \citet{rationalizing-neural-extractive-annotations} tackles text classification by learning the distribution of annotations given the text and that of the target class given the annotations. Interestingly, an additional regularization term is added to the loss to produce annotations that are short and coherent. The model makes use of high-capacity recurrent neural networks \cite{bidirectional-rnn}, thus it is tested on large amounts of training data to prevent overfitting. This work was later refined by \citet{interpret-predictions-differentiable-binary-values}, who proposed a probabilistic version of a similar architecture, where a latent model is responsible for the generation of \textit{discrete} annotations. The main advantage of predicting discrete annotations is that it is possible to constrain their maximum number per sample, thus effectively controlling sparsity. However, it usually requires a large number of data points to be effective. \\
The first to exploit annotations (also called \textit{rationales} in this case) in a low resource scenario were \citet{rational-improve-text-zaidan2007} and \citet{modeling-annotators-generative}, by means of a rationale-constrained SVM \cite{svm} and a probabilistic model. Moreover, the latter is realized as a log-linear classifier that makes heavy use of feature-engineering. On the other hand, when annotations are defined on features rather than on samples, one can use the Generalized Expectation (GE) criteria \cite{druck2007reducing, generalized-expect-criteria} to improve the performance of classifiers. \\ 
Annotations can also be incorporated in the loss function as done in \citet{sequence-classification-human-attention}, where an attention module \cite{attention-is-all-you-need} on top of an LSTM \cite{lstm} is forced to attend words in a document. A similar approach has been successfully applied by \citet{deriving-attention-from-human-raionales} to the weak supervision problem. However, the model assumes one \textit{source domain}, with supervised labels, to learn an attention generation module that is then applied to the target domain. In contrast, our method can be built on a given embedding space with minimum supervision. \\
Apart from incorporating prior knowledge in the form of annotations, we also mention other ways in which neural networks can be augmented: first-order logic \cite{augmenting-nn-first-order-logic,dnn-logic-rules}; a corpora of regular expressions \cite{marrying-regexp-neural-networks}; or massive linguistic constraints \cite{dnn-knowledge-rules}. While generally powerful and effective, all these methods require domain-specific expertise to define the additional features and constraints that have to be explicitly incorporated into the network; our method, instead, is designed to be task-agnostic. In a different manner, the SoPA architecture of \citet{sopa-weighted-finite-state-machines} learns to match surface patterns on text through a differentiable version of finite state machines, which relies on fixed-length and linear-chain patterns to classify a document. 
Instead, BabbleLabble (BL) \cite{training-nl-explanations-2019} is a method for generating weak classifiers from natural language explanations when supervision is scarce. On the one hand, BL works well because it exploits a domain-specific grammar to parse explanations; on the other hand, this grammar must be carefully designed by domain experts. \\ Perhaps the most similar to this work, the Neural Bag Of Words (NBOW) model \cite{nbow} takes an average of the elements belonging to an input sample and applies a logistic regression to classify a document. Its extension, NBOW2 \cite{nbow2}, computes an importance score for each word by comparing it with a single reference vector that is learned. Despite the underlying idea being similar, we propose a more general mechanism to focus on relevant words and use the given annotations. \\
As a final remark, notice that our setting is substantially different from the more common literature on few-shot learning \cite{prototypical-networks-supervised-kmeans,few-shot-gnn,few-shot-learning-closer-look}, where the goal is to classify classes that were \textit{unseen} at training time. Here, we use annotations to be able to associate concepts with the right class, something which must be known in advance for the method to work properly.

In the following, we describe the architecture. As we shall discuss, the model has a strong inductive bias that reflects our intuition about how a model should work in the absence of large amounts of data. Hereinafter, we refer to our new architecture with the name \parcus{} (the Latin word for \quotes{\textit{parsimonious}}).

\section{The \parcus{} model}
\label{model}
Let us consider a classification task in which a very small labelled dataset $\mathcal{D} = \{ (x_1,r_1,y_1),\dots, (x_L, r_L, y_L) \}$ is given. Here, $x_i$ is an input sample, $r_i$ represents the extra (optional) annotations provided by a human, and $y_i$ is a discrete target label.  For the purpose of this paper, an input is a set of tokens $x_i = \{ x^1_i,\dots,x_i^{T_i}\}$ of arbitrary size $T_i$. In addition, $x_i^j \in \mathbb{R}^n$, where $n$ is the size of an embedding space. Finally, each token in the training set may be marked as relevant or not by annotators, \ie $r_i = \{ r^1_i,\dots,r_i^{T_i} \} \in \{0,1\}^{T_i}$.

\subsection{Intuition}
\label{sec:intuition}

When humans are asked to solve a classification problem after seeing a few examples, they tend to look for very simple patterns across the dataset, and text classification is an excellent use case. For instance, assume the word \quotes{excellent} is important to classify a movie review as positive; if we were to work in the character space, a straightforward solution would be to match specific (sub-)strings in the input, an instance of the so-called \textit{pattern matching} technique. At the same time, however, humans are able to generalize to \textit{semantically similar} concepts, and our goal is to exploit similar embeddings to reflect this ability. \\
In this work, we transfer the concept of pattern matching into the embedding space, where semantically similar words are assumed to have similar representations. We achieve this via a mechanism that outputs a probability of matching between an input token and a \textit{prototype vector}, the latter of which is learned to capture discriminative concepts. Differently from bag-of-words methods of Section \ref{related-works}, our model can accommodate multiple prototypes and focus on concepts that are useful for the task. \\ 
Moreover, in order to guide the learning process using the extra annotations, it seems sensible to magnify the error for those tokens that have been marked as relevant by annotators. Notwithstanding the simplicity of the idea, the underlying challenge this work addresses is to effectively embed such human knowledge into the prototypes. In other words, each matching probability should be highly correlated with a particular target class. In addition, it would be desirable that a user could understand what the model has learned, something of great interest when working with the uncertainty caused by a scarce number of training samples. In this respect, we will provide a practical example in Section \ref{sec:explainability}.

\begin{figure*}
     \centering
     \resizebox{0.8\textwidth}{!}{\input{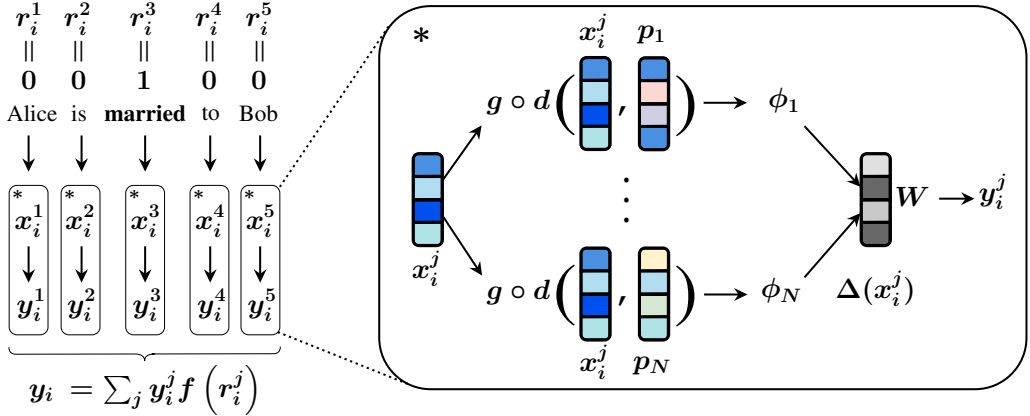}}
     \caption{The \parcus{} architecture is applied to the $i$-th example of a dataset, \ie \quotes{Alice is married to Bob}, with \quotes{married} being highlighted. We extract features by computing the similarity between the token's embedding and the prototypes of our model. Then, we combine these features with a linear layer that outputs per-token predictions. At training time only, predictions are multiplied by a boosting factor $f(r_i^j)$. Individual tokens' predictions are then summed to yield the sentence prediction $y_i$.}
    \label{fig:architecture}
\end{figure*}

We now show how to compute and combine multiple matching probabilities, and then we introduce a technique to incorporate extra annotations in the training process. It is worth mentioning that both techniques have been designed to coexist, even though the latter is not strictly necessary. To provide a graphical representation of the proposed architecture, Figure \ref{fig:architecture} depicts a use case for text classification.

\subsection{Concept Matching}
\label{sec:concept-matching}
We now present the core mechanism that implements concept matching. Let us define a set $\mathcal{P} = \{p_1,\dots,p_N\ \mid p_i \in \mathbb{R}^n\}$ of prototypes to be learned, where $N$ is an hyper-parameter of the model. Each $p_i \in \mathcal{P}$ should ideally adapt to be similar (in the embedding space) to the representation of important tokens.

To learn the $N$ prototypes, we employ the \textit{cosine similarity} metric. Cosine similarity has been often used to measure semantic similarity \cite{cosine-similarity-semantic}; its co-domain ranges from $-1$, \ie opposite in meaning, to 1, \ie same meaning, with 0 indicating uncorrelation. Ideally, we would like our prototypes to have high similarity with the relevant tokens in the input. To this aim, we further define an exponential activation function $g : [-1, 1]\rightarrow [0,1] $ that takes the distance between a token $x_i^j$ and a prototype $p_k$ and outputs a probability of matching:
\begin{align}
P(x_i^j \text{ matches } p_k) = g(d(x_i^j, p_k)) = a^{d(x_i^j, p_k) - 1}
\label{eq:gating}
\end{align}
where $a$ is an hyper-parameter and $d(x_i^j,p_k)$ computes the cosine similarity between $x_i^j$ and $p_k$. In practice, the closer to 1 the similarity is, the greater the output of this gated activation, and $g(v) = 1 \Leftrightarrow v = 1$. By choosing a high value of $a$ we strongly penalize tokens that are associated with low similarity scores.

\subsection{Combining Multiple Prototypes}
\label{sec:combining}
As we saw, Equation \ref{eq:gating} computes the matching probability between a token and a prototype. Because we have $N$ prototypes, we treat the associated $N$ probabilities as a feature vector for the input token, and we denote each feature as $\phi_k(x_i^j) = g(d(x_i^j, p_k)) \ \forall k \in 1,\dots,N$. Interestingly, working with matching probabilities allows us to combine all of them through AND/OR/XOR logical functions. An approximation of such functions can be straightforwardly implemented through the pseudo-differentiable version of $min$ and $max$ \cite{pytorch}, though a fully differentiable version exists:
\begin{align}
&\phi_{AND}(x_i^j) = min(\{\phi_k(x_i^j) \ \ \ \forall k \}) \\
& \phi_{OR}(x_i^j) = max(\{\phi_k(x_i^j) \ \ \ \forall k \})  \\
& \phi_{XOR}(x_i^j) = \phi_{OR}(x_i^j) - \phi_{AND}(x_i^j) 
\label{eq:logic-feat}
\end{align}
In our experiments, the use of $min$ and $max$ functions significantly sped up convergence due to the absence of non-linearities. Moreover, we chose to augment $\Delta(x_i^j)$ with the probability of opposite matching: $\phi_{\neg k}(x_i^j) = g(-d(x_i^j, p_k)) \ \forall k \in 1,\dots,N$. Specifically, when $\phi_{\neg k}(x_i^j) \approx 1$ it means the token $x_i^j$ and $p_k$ have opposite meaning.  Finally, notice how our method differs from NBOW2 \cite{nbow2}, as we use prototypes to compute per-token features rather than an importance score.

\subsection{Inference}
Once we have a feature vector for each token, we need to combine all $F=N+3$ features to output a token prediction $y_i^j$. Let us first define an auxiliary term (omitting the argument $x_i^j$ to make notation less cluttered):
\begin{align}
\Delta(x_i^j) =  [\phi_1,\dots, \phi_N, \phi_{AND}, \phi_{OR}, \phi_{XOR}]
\label{eq:features}
\end{align}
where square brackets denote concatenation. Then, we compute token predictions by linearly combining features:
\begin{align}
y_i^j = \Delta(x_i^j) \mathbf{W} + b
\label{eq:token-prediction}
\end{align}
where $\mathbf{W} \in \mathbb{R}^{F\times C}$ is a matrix of parameters (multi-class prediction with $C$ classes) and $b$ is the (optional) bias. The linear model is useful when we want the user to analyze the importance given to each matching probability, as discussed in Section \ref{sec:intuition}, as well as to restrict the number of parameters of the model (see discussion below). Finally, the input prediction is just a sum of the individual $y_i^j$
\begin{align}
y_i = \sigma(\sum^{T_i}_j y_i^j),
\label{eq:doc-prediction}
\end{align}where $\sigma$ is the \textit{softmax} activation.

\paragraph{Discussion} Regularization of the matrix $\mathbf{W}$ plays an important role to answer our research questions. We use both L1 and L2 regularization terms on $\mathbf{W}$, as done in \cite{elastic-net}, for two main purposes. First, the L1 term enforces sparsity and 
discourages the mixing of too many concepts. Secondly, L2 limits the magnitude of the weights, hence avoiding over-compensation of low cosine similarity scores. Consequently, in order to increase one of the matching probability features, the model is encouraged to make changes to the prototypes rather than to the linear weights; In other words, relevant information for the task will be stored inside prototypes in the form of semantic embeddings. \\

\subsection{Annotation-driven Error Boosting}
\label{sec:error-boosting}
So far, we have not made use of annotations, which are of fundamental importance to guide the learning process in low-resource scenarios. To learn prototypes that match relevant concepts, the proposed technique should weight the importance of tokens rather than whole samples. It follows a boosting approach \cite{adaboost} is not feasible in this scenario; instead, our method exploits prior information in an efficient way. The idea is to modify the error associated with specific tokens to encourage prototypes to be similar to them. To be more precise, \textit{at training time} we modify Equation \ref{eq:doc-prediction} to take into account the given annotations:
\begin{align}
y_i = \sigma(\sum^{T_i}_j y_i^j \cdot f(r_i^j)),
\label{eq:doc-prediction-rationale}
\end{align}
where $f : [0,1] \rightarrow \mathbb{R}$ is an arbitrary exponential function of our choice that boosts the error, \eg $f(r_i^j) = e^{r_i^j}$. In terms of learning, $f(r_i^j)$ boosts the gradient of highlighted tokens while leaving unchanged the rest (\ie if $r_i^j$ is 0, our $f(r_i^j)$ outputs a multiplicative factor of 1). From a mathematical standpoint, we cannot achieve the same result as Equation \ref{eq:doc-prediction-rationale} by means of an additional loss term, as done in \citet{rationalizing-neural-extractive-annotations}, because gradients would be summed and not multiplied as done here.

\subsection{Model complexity and inductive bias}
\label{sec:inductive-bias}
We conclude with remarks on the model complexity. The total number of parameters is $\Theta(Nn + FC)$, which could be much larger than that of a linear model ($\Theta(nC)$) when $N$ is high and $C$ is small. Usually, a restricted number of parameters serves to counteract overfitting by limiting the hypotheses space of the model \cite{vapnik-slt}. However, this work tackles the problem from a novel perspective, as we prevent the prototype weights from \textit{freely} changing. Specifically, prototype weights vary in a way that depends on the given embedding space, because the learning process makes them similar to some token $x_i^j$. If we allowed the weights to freely change, we would get something similar to an MLP; our experiments show how this way of constraining the weights fits particularly well the use case we are considering. Finally, notice that \parcus{} ignores the structural dependencies between input tokens; this is intended, as it is not feasible to learn complex interactions with only a few data samples. Nonetheless, if semantic representations $x_i^j$ are obtained with a pre-trained model, they will usually carry some structural information as well.

\subsection{Theoretical analysis}
The choice behind the concept matching mechanism of Section \ref{sec:concept-matching} is backed up by a theoretical explanation. Indeed, in the limit of the gating parameter $a$, Equation \ref{eq:gating} converges to the discontinuous Kronecker delta function $\delta(x_i^j,p_k)$ that is 1 when its arguments are equal and 0 otherwise; hence, Eq. \ref{eq:gating} is a sound approximation of a \quotes{hard match} function.

\begin{proposition}{}
\label{prop:convergence}
Let $x,y \in \mathbb{R}^n$ and  $d:  \mathbb{R}^n\times\mathbb{R}^n \rightarrow (-\infty, 1]$ be a function such that $d(x,y)=1 \iff x=y$.  Then, the sequence of functions $\{f_a\}_ {a>1}$  with $f_a=a^{d(x, y) - 1} $ is \textit{pointwise convergent} to $\delta(x,y)$.
\end{proposition}
\begin{proof}
To prove pointwise convergence, it is sufficient to show that
\begin{align*}
&\lim_{a\rightarrow+\infty} a^{d(x,y)-1} = \delta(x,y)\ \ \forall x,y.
\end{align*}Because the $d$ cannot take values greater than 1, it follows that $d(x,y)-1\leq 0$, and the equivalence holds if and only if $x=y$. Therefore,  $f_a(x,y) \rightarrow 0 $ for $x\neq y$ and $f_a(x,y) \rightarrow 1$ when $x=y$, that is $\delta(x,y)$. 
\end{proof}
From this proposition we can make another important consideration. Given that it is not possible to have \textit{uniform} convergence to any discontinuous function, some parts of $\delta(x,y)$ will be approximated more easily than others. Specifically for cosine similarity, it can be shown that the area comprised between the two functions, \ie the {error} of our approximation, is $\frac{a^2-1}{a^2\ln a}$; nonetheless, reasonable values of $a$ guarantee good performances and stable learning curves in our experiments. In summary, this result reveals that the best we can do is to look for approximations that satisfy desirable properties, for example being more accurate near the discontinuity and more \quotes{permissive} elsewhere.

\section{Experiments}
\label{experiments}

This section reports the experimental setting as well as our experimental findings. We compare \parcus{} against a large number of baselines. Additionally, we perform an in-depth analysis of our model through ablation studies and qualitative analyses of the effect of some hyper-parameters. Then, we consider a practical scenario in which a user wants to gather insights on how \parcus{} predicts a class for each input sample. We use natural language processing benchmarks to validate our model, and all code to reproduce and extend our experiments is made available\footnote{\url{https://github.com/facebookresearch/parcus}.}.

\subsection{Experimental Setting}
\paragraph{Datasets} We empirically validate our method on two different datasets. First, the \moviereview{} dataset \cite{rational-improve-text-zaidan2007} contains balanced positive and negative movie reviews with annotations. Secondly, we use the highly imbalanced (8\% of positive samples) \spouse{} dataset from \citet{training-nl-explanations-2019}, where the task is to tell whether two entities in a given piece of news are married or not. This is a harder task than standard classification, as the same document can appear multiple times with different given entities and the background context greatly varies. Datasets statistics are reported in Table \ref{tab:dataset-statistics}. We provide annotations for 60 randomly chosen positive samples of the \spouse{} dataset; this process is fast and aims at replicating real world scenarios where labels are scarce and hard to collect.
\begin{table}[ht]
\vskip -0.1in
\caption{Datasets' statistics.}
\label{tab:dataset-statistics}
\begin{center}
\begin{small}
\begin{sc}
\begin{tabular}{llllll}
& \bf Train & \bf Valid. & \bf Test & \bf Annotations 
\\ \hline
\spouse					& 22195 & 2796 & 2697 & 60 \\
\moviereview		& 1800 & - & 200 & 1800
\end{tabular}
\end{sc}
\end{small}
\end{center}
\vskip -0.2in
\end{table}

\begin{table*}[t]
\caption{Hyper-parameters tried during model selection.}
\label{tab:hyper-parameters}
\vskip 0.1in
\begin{center}
\begin{scriptsize}
\begin{sc}
\begin{tabular}{l|c|c|c|c}
    & \textsc{Linear} & \textsc{MLP/NBOW(2)/DAN} & \textsc{BERT+finetune} & \textsc{Ours} \\ \hline
   \textsc{Learning rate} & $\{1e\text{-}2, 1e\text{-}3, 1e\text{-}4\}$ & $\{1e\text{-}3, 1e\text{-}4\}$ & $\{2e\text{-}5, 3e\text{-}5, 5e\text{-}5\}$ & $\{1e\text{-}2\}$ \\
   \textsc{L1} & - & - & - & $\{1e\text{-}2, 1e\text{-}3\}$ \\
   \textsc{L2} & $\{1e\text{-}1, 1e\text{-}2, 1e\text{-}4\}$ & $\{1e\text{-}2, 1e\text{-}4\}$ & - & $\{1e\text{-}3, 1e\text{-}4\}$ \\
   \textsc{Epochs} & $\{50, 100, 150\}$ & $\{100, 500\}$ & $\{2, 4, 10\}$ & $\{500\}$ \\
   \textsc{Hidden units} & - & $\{8, 16, 32 \}$ & - & - \\
   \textsc{Batch size} & 32 & 32 & 8 & 32 \\
   $N$ & - & - & - & $\{5, 10 \}$ \\
   $f(r)$ & - & - & -  & $\{e^r, 5^r, 10^r\}$ \\
   $a$ & - & - & - &  $\{10, 100\}$
\end{tabular}
\end{sc}
\end{scriptsize}
\end{center}
\vskip -0.1in
\end{table*}

\paragraph{Setup} We measure performances on the given test set while varying the number of training data points. We use balanced train splits for all models; on \moviereview, the validation set is taken as big as the training one to simulate a real scenario. As for \spouse, we use the given validation set for model selection to fairly compare with the results of \citet{training-nl-explanations-2019}. We chose the pre-trained (unsupervised) base version of BERT \cite{bert} to provide an embedding space to our method and to other neural baselines. \\
We repeat each experiment 10 times with different random splits; importantly, we train and validate different models on the same data splits. The hyper-parameters for (hold-out) model selection are reported in Table \ref{tab:hyper-parameters}. Moreover, to avoid bad initializations of the final re-training with the selected configuration, we average test performances over 3 training 
runs. The optimized measure is Accuracy for \moviereview{} and F1-score for \spouse. \parcus{} is trained by gradient descent in an end-to-end fashion, from the prototypes to the linear weights. We optimize the Cross-Entropy loss using Adam \cite{adam}.
\paragraph{Methods} To have a good comparison with embedding-based models other than those reported in the literature, we trained a linear model (Linear) and a single-layer MLP, as well as NBOW \cite{nbow}, NBOW2 \cite{nbow2} and the Deep Averaging Network (DAN) of \cite{dan-model}. We also fine-tune BERT using the suggested hyper-parameters \cite{bert}, adding 10 to the possible training epochs. \\ On \spouse, we devised a regular expression that associates specific sub-strings (\textit{\quotes{wife}, \quotes{husb}, \quotes{marr}} and \textit{\quotes{knot}}) to the positive class; ideally, models should be able to focus on such words but also generalize. Moreover, Traditional Supervision (TS)  and Babble Labble (BL-DM) were taken from the work of \citet{training-nl-explanations-2019}: the former method is a logistic regression using n-gram features, whereas the latter is a complex pipeline tested on 30 natural language explanations provided by humans. Notably, BL-DM exploits the relational information of the \spouse{} dataset via task-specific grammar and parser, while \parcus{} simply ignores sentences where the entities of interest are not present. \\
On \moviereview, we also report results of an SVM \cite{rational-improve-text-zaidan2007} and a log-linear model on language features \cite{modeling-annotators-generative}, both of which are specifically designed to exploit additional annotations.\\
Finally, we performed a number of ablation studies to isolate the effect of different techniques:
\begin{enumerate*}[label=(\roman*)] 
\item an MLP with the error boosting technique (\textsc{MLP-W. h.}) to validate the use of prototypes; 
\item our method without highlights (\textsc{\parcus-Wo h.}) to assess the impact of rationales;
\item our method with no logical features (\textsc{\parcus-No-Logic});
\item our method with $\phi_k$ features only (\textsc{\parcus-}$\phi_k$);
\item our method with bilinear rather than cosine similarity (\textsc{\parcus-Bilinear}) to show the importance of constrained weights;
\item Parcus where the input is the average of all input tokens (\textsc{\parcus-Avg});
\item Parcus where centroids are pre-computed using the unsupervised k-means algorithm (\textsc{\parcus-KMeans})
\end{enumerate*}.

\subsection{Results \& Discussion}
Table \ref{tab:results} presents all our empirical results, including the ablation studies. 
\begin{table*}[ht]
\caption{Results for all datasets. Standard deviation is shown in brackets. We report the F1-score as the evaluation metric for \spouse{} and the accuracy for \moviereview.}
\label{tab:results}
\vskip 0.15in
\begin{center}
\begin{scriptsize}
\begin{sc}
\begin{tabular}{l|c|c|c|c|c|c|c|c}
\multicolumn{9}{c}{\textsc{\spouse}} \\ \\
  \textsc{Model/Train Size} & 10 & 30 & 60 & 150 & 300 & 3K & 10K & \\ \hline
   \textsc{Tuned Regexp}     & - & - & - & - & - & - & - & 40.5 \\
    \textsc{TS} &  - & 15.5 & 15.9 & 16.4 & 17.2 & 41.8 & 55.0  \\
   \textsc{BL-DM (30 expl.)}   & - & - & - & - & - & - & - & $\mathbf{46.5}$  \\
   \textsc{Linear}      & 18.2 (1.3) &  20.6 (1.4) & 22.5 (1.4) & 26.1 (1.1) & 26.1 (1.2) & - & - & - \\
   \textsc{MLP}         & 17.9 (2.4) & 20.2 (3.1) & 18.3 (0.6) & 23.3 (1.2) & 24.1 (1.3) & - & - & -  \\
   \textsc{NBOW} & 21.0 (2.3) & 21.8 (1.7) & 24.0 (1.0) & 27.4 (2.0) & 28.2 (1.8) & - & - & -  \\
   \textsc{NBOW2} & 19.5 (2.6) & 22.3 (1.9) & 25.9 (1.4) & 29.6 (1.5) & 31.7 (2.1) & - & - & -  \\
   \textsc{DAN} & 21.8 (3.2) & 24.1 (2.5) & 26.6 (1.7) & 28.2 (1.6) & 29.2 (1.5)  & - & - & - \\
   \textsc{BERT+finetuning} &  16.9 (2.6) & 20.2 (2.1) & 23.4 (1.2) & 32.1 (2.0) & 35.5 (3.2) & - & - & - \\
(Abl.) \textsc{MLP w. h.} &  16.7 (1.4) & 20.8 (2.7) & 20.9 (1.6) & 22.7 (1.8) & 23.1 (2.0) &  - & - & -  \\
   (Abl.) \textsc{\parcus-wo h.} &  27.0 (2.2) & 31.6 (2.5) & 34.2 (2.3) & 41.8 (2.1) & $\mathbf{44.0}$ (1.2) &  - & - & - \\
   (Abl.) \textsc{\parcus-$\phi_k$}  & 32.4 (4.5) & 34.4 (4.2) & 37.8 (2.7) & 42.7 (1.0) & 41.4 (2.4) &  - & - & -  \\ (Abl.) \textsc{\parcus-no-logic}  &  32.7 (3.4) & 34.5 (3.9) & 36.8 (2.6) & 42.7 (1.6) & 42.0 (1.9) &  - & - & -  \\
   (Abl.) \textsc{\parcus-Avg} & 22.9 (3.7) & 26.5 (2.9) & 28.8 (2.2) & 30.5 (1.1) & 32.7 (0.9)  & - & - & -\\
   (Abl.) \textsc{\parcus-KMeans} & 30.3 (2.0) & 33.5 (0.5) & 32.93 (1.0) & 32.8 (0.9) & 34.2 (1.3)  & - & - & - \\
   (Abl.) \textsc{\parcus-Bilinear}  & 29.1 (4.5) & 31.4 (5.9) & 36.0 (5.4)& 36.1 (5.1) & 33.1 (3.0) & - & - & -  \\ \hline
   \textsc{\parcus}                               &  $\mathbf{34.0}$ (4.5) & $\mathbf{36.6}$ (4.3) &  $\mathbf{40.3}$ (2.5) & $\mathbf{43.7}$ (1.7) & 42.9 (1.6) &  - & - & -  \\ \\
 \end{tabular}
\begin{tabular}{l|c|c|c|c|c}
\multicolumn{6}{c}{\textsc{\moviereview}} \\ \\
    \textsc{Model/Train Size} & 10 & 20 & 50 & 100 & 200 \\ \hline
   \textsc{SVM + rationales}       & - & 65.4 & - & 75 & 83.2  \\
   \textsc{Log-linear + rationales}       & - & 65.8 & - & 76 & 83.8 \\
   \textsc{Linear}   & 60.4 (3.4) & 64.0 (3.5) & 70.2 (2.0) & 77.2 (2.6) & 80.3 (3.1) \\
   \textsc{MLP}   & 59.1 (4.1) & 62.6 (4.2) & 69.7 (2.4) & 73.3 (3.8) & 80.0 (3.0) \\
   \textsc{NBOW}   & 62.6 (4.6) & 65.7 (4.8) & 73.9 (1.6) & 78.0 (2.0) & 81.2 (3.6) \\
   \textsc{NBOW2}   & 61.5 (4.5) & 64.3 (4.9) & 72.9 (1.4) & 78.9 (4.4) & 83.6 (1.8) \\
   \textsc{DAN} & 61.5 (6.2) & 62.3 (4.8) & 72.9 (3.3) & 78.7 (3.2) & 82.35 (2.7) \\
   \textsc{BERT+finetuning} & 53.5 (2.0) & 54.8 (4.9) & 59.7 (4.5) & 67.7 (4.3) & 79.2 (2.5) \\
   (Abl.) \textsc{MLP w. h.}  & 61.5 (4.6) & 63.1 (5.8) & 68.9 (7.0) & 72.4 (8.5) & 74.6 (5.8) \\
   (Abl.) \textsc{\parcus-wo h.}  & 61.2 (4.3) & 64.9 (5.0) & 74.3 (2.4) & 78.6 (2.3) & $\mathbf{84.6}$ (2.8) \\
   (Abl.) \textsc{\parcus-$\phi_k$}  & 66.1 (5.7) & 68.4 (3.5) & $\mathbf{77.8}$ (2.0) & 80.7 (3.0) & 83.4 (2.4) \\
   (Abl.) \textsc{\parcus-no-logic}   & 66.9 (5.9) & 67.9 (3.5) & 75.5 (4.0) & $\mathbf{81.0}$ (2.4) & 83.7 (2.7) \\ 
   (Abl.) \textsc{\parcus-Avg} & 62.1 (4.9) & 62.5 (4.4) & 71.0 (3.5) & 73.3 (3.1) & 79.0 (3.4) \\
   (Abl.) \textsc{\parcus-KMeans} & 54.4(5.0) & 53.2 (3.4) & 54.2 (2.6) & 53.6 (2.7) & 58.0 (2.4) \\
   (Abl.) \textsc{\parcus-Bilinear}  & 57.5 (5.1) & 61.9 (6.7) & 70.4 (3.7) & 75.3 (2.9) & 78.3 (3.6) \\ \hline
   \textsc{\parcus}        & $\mathbf{67.2}$ (5.5) &  $\mathbf{70.1}$ (5.6) & $76.6$ (2.4) & $80.0$ (2.6) & $83.8$ (2.8) \\ \\
 \end{tabular}
\end{sc}
\end{scriptsize}
\end{center}
\vskip -0.1in
\end{table*}     
Results highlight that \parcus{} has strong performances in a low data regime, validating intuition and theoretical results. On \spouse, our model strongly outperforms other neural baselines and reaches the \textit{manually tuned} regular expression with just 60 training points. Moreover, TS needs $\approx$\textit{50x} more data to achieve similar performance. We also found that TS performs much worse than our linear baseline (hence the need for a fair comparison in the embedding space). Surprisingly, only 10 data points are sufficient to perform better than almost all baselines with a training size of 300, a \textit{$>$30x} improvement which does not depend on the chosen embedding space. With 300 datapoints and no annotations, our model has an average F1 score very close to that of BL-DM. Notice that the reported result (BL-DM, 46.5) is not averaged over multiple runs, and one of our random splits achieves a test score of 46.3; this indicates the need for robust evaluation when it comes to experimenting with few data points/natural language explanations. Overall, we found that the proposed approach can be helpful when data is greatly imbalanced and diverse in nature, and outperforms powerful models like BERT that are quite performing when fine-tuned on relatively small datasets \cite{bert,ulmfit}. \\
Similar arguments apply to \moviereview, where our model improves over the baselines. Interestingly, \parcus{} is able to improve the state of the art by a large margin when very few data points are used. Here, NBOW and NBOW2 models proved to be the strongest competitors, as they rely on the mean representation of a document. \\ Overall, the gap is more evident as training size is very scarce, even when compared to other baselines that use extra annotations.  This suggests the model could be a good fit for all those practical scenarios where the data gathering process is just started and one wants to boost performances by means of extra annotations.

\subsection{Ablation Studies}
We performed ablation studies on both datasets to understand whether the improvements are only due to prototypes, error boosting technique or both. Overall, we observe that the use of prototypes provides a consistent improvement with respect to the other baseline, and this is especially evident on the \spouse{} dataset. Interestingly, \textsc{MLP w. h.} does not benefit from error boosting, which is in accord with the fact that unconstrained weights make it more difficult to select and isolate the contribution of relevant tokens. In addition, it seems that the logical and opposite matching features can help to boost the average performance, as \textsc{\parcus-$\phi_k$} and \textsc{\parcus-no-logic} always perform worse than \parcus{} on \spouse{}. Because annotations guide the learning process, these are most important in the extremely low resource scenario, but their effect slowly fades as the training size increases; contrarily to our expectations, \parcus{} performs even better on larger amounts of training points without annotations. This indicates that, at a certain point, annotations may regularize the model too much, and it suggests future works on \textit{adaptive} error boosting functions. Finally, note that neither averaging tokens nor pre-computing centroids seem beneficial; indeed, models like DAN better exploit the average using an MLP on top of the averaged representation, while we force the model to align to some relevant input token. Also, the use of pre-computed centroids will make the model focus on the most common semantics in the dataset, which are not necessarily the most adequate to solve the task.

\subsubsection{More general distance functions}
In Section \ref{sec:inductive-bias} and in the above discussion, we argued that the inductive bias of our architecture is favorable for the specific problem we are tackling. Here, we empirically validate our statement by showing that the use of a more general distance function $d$ tends to overfit the data and achieves significantly worse performances. In particular, we substitute the cosine similarity with its bilinear counterpart $d_{W_b}(x,p) = \tanh(x^TW_bp)$, where $W_b \in \mathbb{R}^d \times \mathbb{R}^d$, and we ran the experiments on \spouse{} and \moviereview{} (shown in Table \ref{tab:results} as \textsc{\parcus-BILINEAR}). Bilinear similarity can be seen as a generalization of cosine similarity when individual features are given different importance (specified by the matrix $W_b$). However, the number of parameters is quadratic in the dimension of the given embeddings, and this matrix is unconstrained, unlike prototypes. \\ Overall, we observe that the use of bilinear similarity still yields good performances on the \spouse{} task, but it is not capable of generalizing well on \moviereview{} where the average number of tokens in each sentence is much higher. The reason may be that since \spouse{} contains pieces of news related to different topics, focusing solely on those concepts related to marriage may help. \\
These empirical results reinforce the belief that constraining the weights to match specific concepts in a low-resource scenario helps to generalize to new instances.
\subsection{Qualitative analysis on the effect of $a$}
\label{sec:qualitative-analysis}
\begin{figure}[t]
     \centering
     \includegraphics[width=0.4\textwidth]{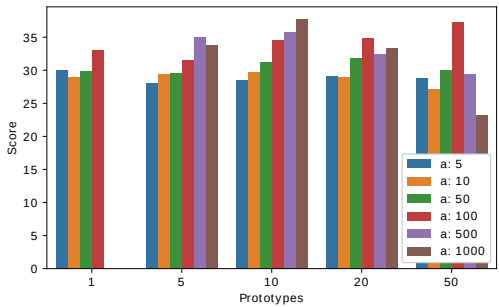}
     \caption{Here, 60 data points are used to train \parcus{} on the \spouse{} dataset. Our analysis reveals how larger values of $a$ should be associated with a reasonable number of prototypes (much less than the size of the training set) to achieve good performances.}
     \label{fig:prototype-vs-gate}
\end{figure}
The parameter $a$ plays an important role in controlling how strict the model is in considering a matching to be highly probable. Larger values of $a$ should produce prototypes that are more specific to a single concept, while smaller values (but still greater than 1, see Proposition \ref{prop:convergence}) allow a prototype to match less similar tokens. To further confirm our intuition, we run an experiment on the \spouse{} dataset where we analyzed the trade-off between the value of $a$ and the number of prototypes. Figure \ref{fig:prototype-vs-gate} shows our results for 60 data points. We immediately see that using just 1 prototype with a large value of $a$ may be too restrictive to solve the task, which is in accord with common sense. However, the general trend we observe is that enforcing separation of concepts is usually beneficial, provided the number of prototypes is sufficiently high.
\subsection{Gaining insights from the learned weights}
\label{sec:explainability}
\begin{figure}[t]
     \centering
     \includegraphics[width=0.5\textwidth]{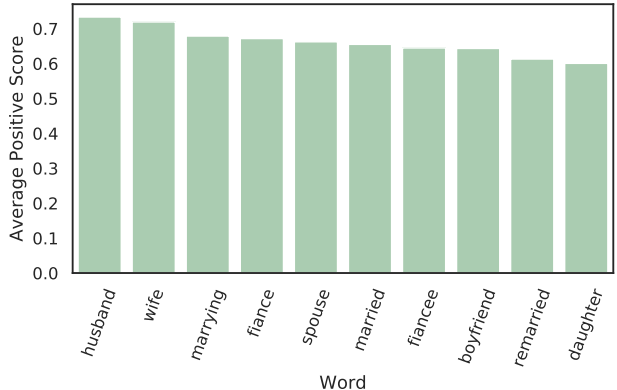}
     \caption{Top-10 most relevant tokens for positive prediction, averaged on \textit{unseen} data.}
     \label{fig:ranking-predictions}
\end{figure}

The learned weights of the proposed model can be inspected to gain insights on what concepts it focuses on and how they are related. To show this, we train a model using $N$=3 prototypes on 60 examples taken from the \spouse{} dataset. Then, we rank the tokens' outputs of sentences belonging to unseen data, so that the outputs with the highest rank correspond to semantic concepts that have been considered relevant for the task by the model. Specifically, as shown in Figure \ref{fig:ranking-predictions}, the model learns to focus on words related to marriage, as well as syntactic variations associated with similar semantics. Importantly, some of the words were not given as annotations in the training set, meaning that the model is also able to recognize similar concepts. \\
We additionally show that annotators' knowledge has been effectively incorporated into the prototypes, and how the features of Equation \ref{eq:features} have been combined together. We start by inspecting the magnitude of the linear weights $\mathbf{W} \in R^{F\times2}$; specifically, if the $i$-th feature is discriminative for a class $c$, then the $i$-th row of $\mathbf{W}$ will have the $c$-th element larger than the others. In our example, we find that $\phi_1$ was important for positive predictions, whereas the other features did not contribute much to a particular class. We then perform top-10 cosine similarity ranking between tokens and the prototype $p_1$. From the most similar to the least one, we obtain: \textit{husband}; \textit{marriage}; \textit{marrying}; \textit{wife}; \textit{married}; \textit{marry}; \textit{fiance}; \textit{wedding}; \textit{fiancee}; and \textit{girlfriend}. This result gives insights on how \parcus{} has learned to match concepts similar to those provided in natural language form by BL-DM (see Appendix of \citet{training-nl-explanations-2019}).

\section{Limitations and future works}
\label{sec:limitations}
Though \parcus{} performs very well and its learning dynamics follow our intuition, there are some inherent limitations to the method. The first is that it is not possible to uniformly approximate the Kronecher delta function of Proposition \ref{prop:convergence}, and as such we can only study further approximations that work better around the discontinuity. The second is the need to map the input into embedding space before training, which can be restrictive for less common application domains. This has an impact on how easily we can inspect the weights as done in Section \ref{sec:explainability}; however, all domains for which a pre-trained method exists should benefit from our technique.
Also, notice that cosine similarity is just one of the functions that can be used: if we are interested in the magnitude of the vectors when computing similarities, a normalized Euclidean norm can be a valid choice. Interesting future works will be the investigation of \parcus{} performance on larger training sets and its extension to an adaptive version of the error boosting function $f$.
\section{Conclusions}
\label{conclusion}
In this work, we presented \parcus, a new representation learning methodology to perform classification in the low data regime. We coupled matching probabilities with error boosting to focus on concepts that are important for the task at hand. After comparing it with a large number of baselines, the model performed very well and outperformed most of them. We provided theoretical insights on the design of our matching technique, and we make an in-depth analysis of some characteristics of the model as well as many ablation studies. Moreover, we showed with a practical example that the weights can be inspected to see what concepts the model focuses on. In summary, our model can be very useful in tasks where gathering data is challenging, and it can be used to assist users in training a classifier for a very specific task.

\newpage

\bibliography{main}
\bibliographystyle{icml2020}



\end{document}